\pdfoutput=1

\documentclass[11pt]{article}

\usepackage{EACL2023}

\usepackage{times}
\usepackage{latexsym}
\usepackage{makecell}
\usepackage{multirow}
\usepackage{amsmath}

\usepackage{enumitem}
\setlist{nosep}

\usepackage[T1]{fontenc}

\usepackage[utf8]{inputenc}

\usepackage{microtype}

\usepackage{inconsolata}

\title{\emph{[Lions: 1] and [Tigers: 2] and [Bears: 3], Oh My!}\\ Literary Coreference Annotation with LLMs}

\author{Rebecca M. M. Hicke  \\
  Department of Computer Science \\
  Cornell University \\
  \texttt{rmh327@cornell.edu} \\\And
  David Mimno \\
  Department of Information Science \\
  Cornell University \\
  \texttt{mimno@cornell.edu} \\}

\begin{document}
\maketitle
\begin{abstract}
Coreference annotation and resolution is a vital component of computational literary studies. However, it has previously been difficult to build high quality systems for fiction.
Coreference requires complicated structured outputs, and literary text involves subtle inferences and highly varied language.
New language-model-based seq2seq systems present the opportunity to solve both these problems by learning to directly generate a copy of an input sentence with markdown-like annotations.
We create, evaluate, and release several trained models for coreference, as well as a workflow for training new models.
\end{abstract}

\section{Introduction}

Coreference annotation and entity recognition are key tasks for performing a wide variety of textual analyses. They provide important information about texts as well as serving as the foundation for many more complicated forms of analysis. Particularly within the digital humanities (DH), these tasks are often essential for performing large-scale studies of corpora (e.g.\ \citet{underwood2018transformation, papalampidi-etal-2019-movie, brahman-chaturvedi-2020-modeling}).
However, coreference annotation is considerably more difficult than many binary classification tasks.
First, coreference requires nuanced understanding of text, which has been beyond the capabilities of previous NLP.
Second, coreference requires structured output, such as marking spans for entity mentions and coreferent mentions, which has previously required custom software.

Generative large language models (LLMs) have recently demonstrated a capacity to solve both problems \cite{bohnet-etal-2023-coreference,zhang-etal-2023-seq2seq}.
By leveraging massive pretraining collections and billions of parameters, they can identify the subtle, nuanced patterns of language.
In addition, they can generate text that matches specific text markup formats.
This capability suggests that non-expert users may be able to use ``out of the box'' LLMs to generate complicated marked-up text simply by providing examples of the desired input and output.
While we evaluate this process by comparing with existing custom-built coreference systems, we emphasize that the potential impact of this process extends to a much broader class of markup.

To explore the promise of fine-tuning generative LLMs for coreference annotation, we evaluate the capabilities of several models to perform coreference annotation on sentences extracted from literary texts. Previous research has shown that literary texts have unique characteristics \cite{bamman-etal-2020-annotated} that make it difficult to adapt generalized NLP models to literary settings. \citet{zhang-etal-2023-seq2seq} achieve high performance on the LitBank corpus when data from the corpus is included in the fine-tuning dataset; we seek to further explore the capabilities of a model adapted specifically for literary coreference.

In this work, we find that a fine-tuned \texttt{t5-3b} model significantly outperforms a state-of-the-art neural model for literary coreference annotation \cite{otmazgin-etal-2022-f}. In addition, we speculate on the ability of these models to perform more complicated, abstract annotation tasks (e.g.\ identifying character relationships) given its performance on this task.

Specifically, in this work we contribute:
\begin{itemize}
    \item A high-performing fine-tuned LLM and supporting code that can be used to perform coreference annotation on literary data.\footnote{\url{https://huggingface.co/rmmhicke/t5-literary-coreference}}
    \item An analysis of which LLMs are best suited as foundation models for coreference annotation.
    \item An examination of these models strengths and weaknesses for coreference annotation.
\end{itemize}

\section{Related Work}

Many researchers have used neural networks \cite{lee-etal-2017-end, clark-manning-2016-deep, ijcai2019p687} or encoder-only transformer models like the BERT models as the basis for coreference systems \cite{ye-etal-2020-coreferential, joshi-etal-2019-bert, otmazgin-etal-2022-f, wu-etal-2020-corefqa}. These methods are multi-step and perform entity recognition and coreference annotation separately. Some studies have explored using generative LLMs for coreference, but they generally either fine-tune on auxiliary tasks \cite{mullick-etal-2023-better} or use zero- or few-shot prompting \cite{le-etal-2022-shot, 2305.14489}.

\citet{bohnet-etal-2023-coreference} fine-tune two sizes of mT5 (\texttt{xl} and \texttt{xxl}) to output coreference annotations for multi-lingual data.
They annotate speaker interactions fed to the model one sentence at a time. The model outputs either link or append actions, which are used to annotate the coreference clusters in the next model input. Similarly, \citet{zhang-etal-2023-seq2seq} find that seq2seq models such as T5 perform well when directly fine-tuned to output sentences annotated for coreference in a format similar to markdown.
Like \citet{zhang-etal-2023-seq2seq} and unlike \citet{bohnet-etal-2023-coreference}, we fine-tune a model to directly produce inline coreference annotations. Unlike both papers, we do not attempt to link annotations between sentences. We also focus specifically on literary coreference annotation, which \citet{zhang-etal-2023-seq2seq} include but do not foreground, and compare encoder-decoder models to decoder-only models. Finally, we perform a qualitative examination of the fine-tuned models' strengths and weaknesses.

Coreference annotation has been applied to a wide variety of domains, such as movie screenplays \cite{baruah-narayanan-2023-character}, biomedical journals \cite{cohen2017coreference}, and fiction \cite{bamman-etal-2020-annotated}. Coreference annotation for literary texts in a variety of languages has also received a great deal of attention \cite{poot-van-cranenburgh-2020-benchmark, schroder-etal-2021-neural, han-etal-2021-fantasycoref, krug-etal-2015-rule, roesiger-etal-2018-towards}. However, to our knowledge no work has yet focused on fine-tuning and evaluating generative LLMs specifically for literary coreference.

\section{Data \& Methods}

Our training data is drawn from the LitBank corpus \cite{bamman-etal-2019-annotated}, which includes 100 novels written in English before 1923 representing a mix of ``high literary style... and popular pulp fiction'' \cite[p.~2139]{bamman-etal-2019-annotated}. The mixture of publication dates and styles included in the corpus means that we are able to train and evaluate models for a variety of sentence styles. Human coreference annotations are available for the first \textasciitilde2,000 tokens of each text for people, natural locations, built facilities, geo-political entities, organizations, and vehicles \cite{bamman-etal-2020-annotated}.

\begin{table}[]
\small
\caption{Example of an input-output pair used during fine-tuning. In the output, entities are surrounded by brackets and the association cluster is labeled as an integer.}
\centering
\begin{tabular}{p{0.45\linewidth} | p{0.45\linewidth}}
\hline
\textbf{Input} &  \textbf{Output} \\
\hline
Carl thrust his hands into his pockets, lowered his head, and darted up the street against the north wind. & [Carl: 1] thrust [his: 1] hands into [his: 1] pockets, lowered [his: 1] head, and darted up [the street: 2] against the north wind.\\
\hline
\end{tabular}
\label{tab:inputOutput}
\end{table} 

We created a subset of the LitBank corpus containing coreference-annotated sentences from the 92 novels with at least 50 annotated sentences. We standardized the formatting of each sentence by hand in an attempt to regularize punctuation. Then, we created an input and output version of each sample (see Table \ref{tab:inputOutput}) where the input is the plain sentence and the output contains formatted coreference annotations. These were used to fine-tune and evaluate each model.

We withhold five novels entirely from the training dataset and include all sentences (at least 50) drawn from these novels in the test set. From each of the remaining 87 novels, we include 40 sentences in the training dataset, 2 sentences in the validation set, and the remaining sentences (at least 8) in the test dataset. The final dataset had 3,480 sentences for training, 174 sentences for validation, and 4,560 sentences for testing.

We then fine-tuned different sizes of three LLMs to perform literary coreference annotation: four sizes of T5 (\texttt{small}, \texttt{base}, \texttt{large}, \texttt{3b}) \cite{raffel2019}, three sizes of mT5 (\texttt{small}, \texttt{base}, \texttt{large}) \cite{xue-etal-2021-mT5}, and five sizes of Pythia (\texttt{70m}, \texttt{160m}, \texttt{410m}, \texttt{1b}, \texttt{1.4b}) \cite{biderman2023pythia}. mT5 is included to inform future research on multilingual coreference. Because we are interested in supporting users with low access to hardware accelerators, models are included only if they can be fine-tuned on a single GPU. The parameters used for fine-tuning can be found in Appendix \ref{sec:appendixFine}.

Finally, as a baseline we evaluate three spaCy-based coreference annotation systems: \texttt{fastcoref} \cite{otmazgin-etal-2022-f} using the \texttt{LingMess} model \cite{otmazgin-etal-2023-lingmess}, huggingface's \texttt{neuralcoref} (which is based on \citet{clark-manning-2016-deep}), and Explosion AI's \texttt{coreferee}. We do not include the performance of the BookNLP system, the most-used tool for literary coreference annotation, as it is also trained on LitBank and has likely seen some of our test sentences. However, we include a comparison of BookNLP and our fine-tuned \texttt{t5-3b} model's performance on two books not in LitBank: Virginia Woolf's \textit{Orlando} and Radclyffe Hall's \textit{The Well of Loneliness}, which entered the public domain in 2024.

\section{Results}

\begin{table}[t]
\small
\caption{Results for entity recognition and coreference. T5 has the best performance, particularly the 3B scale. FastCoref is a non-seq2seq baseline. The multilingual mT5 model is similar but not as good as T5, while the decoder-only Pythia family fails to add any annotations, correctly repeating only inputs with no annotations.}
\centering
\begin{tabular}{lrrrr}
\textbf{Model} & \textbf{Ent. F1} & \textbf{\Gape[0pt][2pt]{\makecell[c]{Coref.\\F1}}} & \textbf{\Gape[0pt][2pt]{\makecell[c]{Average\\Edit \\ Distance}}} & \textbf{\Gape[0pt][2pt]{\makecell[c]{Exact\\String \\Match}}}\\
\hline
\multicolumn{5}{c}{\textbf{Baselines}} \\
\hline
\texttt{fastcoref} & 50.86 & 40.46 & --- & --- \\
\texttt{neuralcoref} & 41.30 & 29.68 & --- & --- \\
\texttt{coreferee} & 35.04 & 2.81 & --- & --- \\
\hline
\multicolumn{5}{c}{\textbf{T5}} \\
\hline
\texttt{t5-3b} & \textbf{91.03} & \textbf{80.16} & \textbf{0.1} & \textbf{70.72} \\
\texttt{t5-large} & 85.37 & 71.81 & 0.44 & 60.42 \\
\texttt{t5-base} & 83.74 & 61.35 & 1.74 & 47.76 \\
\texttt{t5-small} & 58.01 & 35.96 & 7.36 & 26.05 \\
\hline
\multicolumn{5}{c}{\textbf{mT5}} \\
\hline
\texttt{mT5-large} & 81.90 & 58.78 & 1.77 & 42.39 \\
\texttt{mT5-base} & 70.14 & 47.70 & 5.12 & 15.81 \\
\texttt{mT5-small} & 0.41 & 0.24 & 149.79 & 0.0 \\
\hline
\multicolumn{5}{c}{\textbf{Pythia}} \\
\hline
\texttt{pythia-1.4b} & 0.0 & 0.0 & 1077.79 & 9.06 \\
\texttt{pythia-1b} & 0.0 & 0.0 & 789.7 & 9.04 \\
\texttt{pythia-410m} & 0.0 & 0.0 & 1492.55 & 8.68 \\
\texttt{pythia-160m} & 0.0 & 0.0 & 1617.35 & 7.89 \\
\texttt{pythia-70m} & 0.0 & 0.0 & 1054.16 & 6.91 \\
\hline
\end{tabular}
\label{tab:modelAcc}
\end{table} 

\begin{table*}[t]
\small
\caption{Model results given in common coreference metrics (F1 scores). All Pythia models produce 0.0 F1 in all cases.}
\centering
\begin{tabular}{lccccccc}
\textbf{Model} & \textbf{MUC} & $\text{\textbf{B}}^3$ & $\text{\textbf{CEAF}}_m$ & $\text{\textbf{CEAF}}_e$ & \textbf{BLANC} & \textbf{LEA} & \textbf{CoNLL avg.}\\
\hline
\multicolumn{8}{c}{\textbf{Baselines}} \\
\hline
\texttt{fastcoref} & 80.08 & 56.72 & 58.31 & 38.38 & 56.90 & 54.00 & 58.39 \\
\texttt{neuralcoref} & 52.64 & 36.08 & 39.03 & 24.71 & 32.35 & 31.25 & 37.81 \\
\texttt{coreferee} & 41.49 & 29.77 & 33.76 & 22.09 & 25.18 & 23.46 & 31.12 \\
\hline
\multicolumn{8}{c}{\textbf{T5}} \\
\hline
\texttt{t5-3b} & \textbf{89.19} & \textbf{89.21} & \textbf{89.20} & \textbf{87.20} & \textbf{86.29} & \textbf{85.23} & \textbf{88.53} \\
\texttt{t5-large} & 82.14 & 83.71 & 83.41 & 81.19 & 77.81 & 77.74 & 82.35 \\
\texttt{t5-base} & 71.71 & 77.73 & 75.47 & 72.82 & 70.76 & 65.22 & 74.09 \\
\texttt{t5-small} & 45.62 & 55.82 & 52.03 & 48.10 & 41.97 & 36.47 & 49.85 \\
\hline
\multicolumn{8}{c}{\textbf{mT5}} \\
\hline
\texttt{mT5-large} & 68.26 & 74.75 & 72.72 & 69.98 & 67.07 & 61.37 & 71.00 \\
\texttt{mT5-base} & 61.38 & 64.26 & 62.14 & 56.83 & 55.91 & 49.41 & 60.82 \\
\texttt{mT5-small} & 0.08 & 0.40 & 0.56 & 0.60 & 0.02 & 0.22 & 0.36 \\
\hline
\end{tabular}
\label{tab:corefMets}
\end{table*}



One advantage of seq2seq LLMs is their ability to produce complicated, structured output \textit{as text} without the need for complex structured prediction model architectures.
This means that we can use ``off the shelf'' transformers and fine-tune them using standard methods to produce coreference annotations.
The problem with directly generating marked-up text, however, is that the generated output might not be purely additive: it may change the words in addition to adding annotations.

We therefore evaluate each fine-tuned LLM using four metrics. We measure the fidelity of the output with average Levenshtein distance between the input sentence and the model output stripped of all coreference annotation. We measure annotation accuracy using F1 scores for entity recognition and coreference annotation. Finally, we record whether there is an exact string match between the human-annotated output and the model output (not including leading or trailing spaces). This metric is measured as a percentage of sentences instead of a percentage of entities, as the F1 scores are.

We expect that the entity and coreference F1 scores will be an underestimate of the true model performance. This is for several reasons. The first is that we only count entities as labeled when the cleaned model output (stripped of entity and coreference annotation) is exactly the same length as the input string. Additionally, we only count exact entity or coreference matches. Thus, a match is not made when an entity is selected but misspelled (e.g.\ \texttt{[Helene's:\ 1]} is produced instead of \texttt{[Helen's:\ 1]}) or when a different substring is selected to represent an entity (e.g.\ \texttt{[the study behind the dining-room:\ 1]} is selected instead of \texttt{[the study:\ 1]}). The metric also only counts a coreference annotation as correct if the exact same index is used to identify the cluster. Therefore, if an extra entity cluster is labeled or missed (e.g.\ a sentence is annotated \texttt{[The lady:\ 1] in the room picked up [his:\ 2] hat.} instead of \texttt{[The lady:\ 1] in [the room:\ 2] picked up [his:\ 3] hat.}) some annotations may not be counted even though they are technically correctly identified. Finally, there are cases where the annotation of an entity is somewhat subjective and human observers may side with the initial annotations or the model output. For example, the LitBank annotation for the sentence \texttt{As it chanced, [Dale: 1] lay face down upon the floor of the loft} does not mark ``the loft'' as an entity. However, the model output does, and one could argue that this is a location which should be annotated. For these reasons, we expect that the true performance of these models in the eyes of a human evaluator would be higher than it appears given the strict F1 scores reported. 

In order to provide more generous accuracy metrics that are comparable across other studies, we also use the \texttt{corefud-scorer} developed by Michal Novák and Martin Popel to report the models' performance using seven common coreference evaluation metrics (Table \ref{tab:corefMets}). We count singletons and require exact entity matches. Again, for these calculations we only count an entity as labeled if the clean model output is the same length as the input sentence. However, the scorer simplifies cases in which the same entity has been marked twice --- transforming \texttt{[[he:\ 1]:\ 2]} to \texttt{[he:\ 2]} --- and does not require exact spelling matches between labeled entities.  

\paragraph{T5} Of the models tested, fine-tuned \texttt{t5-3b} achieves the highest performance (Table \ref{tab:modelAcc}). It exactly reproduces 70.72\% of the human-annotated outputs and has F1 scores of 91.03\% for entity recognition and 80.16\% for coreference annotation. Overall, the T5 models outperform all other model families; Pythia is unable to correctly identify any entities or coreference clusters, each mT5 model underperforms the equivalent T5 model, and all baselines are outperformed by all T5 models except \texttt{t5-small} (which is outperformed on coreference by \texttt{fastcoref}).

\begin{table}[]
\small
\caption{Example of less-successful input-output pairs produced from fine-tuned models. The first was produced by \texttt{t5-large}, the second by \texttt{pythia-1b}.}
\centering
\begin{tabular}{p{0.45\linewidth} | p{0.45\linewidth}}
\hline
\textbf{Input} &  \textbf{Output} \\
\hline
He shivered as if he had cold slimy water next his skin. & He shivered as if he had cold slimy water next \textbf{to} his skin. \\
\hline
We must go to Athens. & We must go to Athens. go to [Athens: 2]. go to [Athens: 2]: 2]. go to [A [Athens: 2]:]: 2]: 2 to [A [A [A: 2]: 2]: 2]: 2 to [: 2 [A: 2]: 2]: 2 2 2 2 2 to [: 2 [: 2]: 2]: 2 2 2 2 to [: 2 2]: 2 2...\\

\hline
\end{tabular}
\label{tab:inputOutputResults}
\end{table} 

Larger models do better. The smaller T5 models, particularly \texttt{t5-small}, struggle to accurately match brackets and parentheses. They also frequently miss nested entities such as \texttt{[[her:\ 2] father:\ 1]}, randomly neglect to annotate any entities in a sentence comparable to those for which it has relatively high performance, or repeat substrings and brackets at the end of its output. \texttt{t5-large} sometimes adds extra words to sentences, often when grammatically intuitive (Row 1, Table \ref{tab:inputOutputResults}), replicates only substrings of the original input, or makes other small formatting errors. It also continues to struggle with identifying some of the more complicated multi-word entities and nested entities. Finally, the replication errors for \texttt{t5-3b} are mostly formatting errors or the addition / exclusion of single words or small substrings. The output of this model sometimes still includes hallucinated repetitions, but it is very rare. Most of the annotation ``errors'' made by this model could be judgment calls, or cases where the original annotator had more context. Even this largest model occasionally struggles with matching brackets and nested entities, but this is also extremely rare.

If we examine single word replacements made by \texttt{t5-small} --- for cases when the cleaned output is the exact length of the input --- we find that it struggles with complicated and unusual words (e.g.\ \textit{bordighera}, \textit{schiaparelli}), names (e.g.\ \textit{Katharine} is replaced by \textit{Catarine}, \textit{explained}, and \textit{Katarine}), gender (\textit{Mr.} is replaced by \textit{Mrs.} five times), pronouns (\textit{their} is replaced 82 times by 18 unique strings), and language (\textit{however} is replaced by \textit{nevertheless}, \textit{allerdings}, \textit{cepedant}, and \textit{totuşi} and \textit{Winterbourne} becomes \textit{Hierbourne}). \texttt{t5-base} and \texttt{large} make similar single word replacements, but the translation errors are reduced to changing entities to their spelling in their original language (e.g.\ \textit{Munich} becomes \textit{München}). The replication errors made by \texttt{t5-3b} are almost all misspellings or changes in the plurality of words. Names also continue to confound the model as do parts of speech.

\begin{table}
\small
\caption{Sentences with coreference annotation from fine-tuned \texttt{t5-3b} and BookNLP. The sentences are drawn from Virginia Woolf's \textit{Orlando} (Rows 1 and 2) and Radclyffe Hall's \textit{The Well of Loneliness} (Rows 3--5).}
\centering
\begin{tabular}{p{0.45\linewidth} | p{0.45\linewidth}}
\hline
\textbf{t5-3b} &  \textbf{BookNLP} \\
\hline
Rows of chairs with all their velvets faded stood ranged against the wall holding their arms out... & Rows of chairs with all [their:\ 5] velvets faded stood ranged against the wall holding [their:\ 5] arms out... \\
\hline
[Fathers:\ 1] instructed [[their:\ 1] sons:\ 2], [mothers:\ 3] [[their:\ 3] daughters:
4]. & [Fathers:\ 1] instructed [[their:\ 1] sons:\ 2], [mothers:\ 3] [[their:\ 1] daughters:\ 4]. \\
\hline
... sat [Stephen:\ 1] with [her:\ 1] feet stretched out to the fire and [her:\ 1] hands thrust in [her:\ 1] jacket pockets. & ...sat [Stephen:\ 1] with [her:\ 2] feet stretched out to the fire and [her:\ 2] hands thrust in [her:\ 2] jacket pockets. \\
\hline
[Mrs. Williams:\ 1] glanced apologetically at [her:\ 2]:\ 'Excuse 'im, [Miss Stephen:\ 2], 'e's gettin' rather childish. & [Mrs. Williams:\ 2] glanced apologetically at [her:\ 2]:\' Excuse' [i:\ 1]m, [Miss Stephen:\ 3],' [e:\ 4]'s gettin' rather childish. \\
\hline
When one's getting on in years, one gets set in one's ways, and [my:\ 1] ways fit in very well with [Morton:\ 2]. & When [one:\ 1]'s getting on in years, [one:\ 1] gets set in [one:\ 1]'s ways, and [my:\ 2] ways fit in very well with [Morton:\ 3]. \\
\hline
\end{tabular}
\label{tab:bookNLPComp}
\end{table} 

The CoNLL average coreference score achieved by the fine-tuned \texttt{t5-3b} model exceeds that of BookNLP by 9.5\%; however, the BookNLP system simultaneously provides coreference annotations for each $\sim$2,000 word section of novel at once, whereas the T5 model runs on individual sentences. In order to further compare the two systems, we thus ran both on 100 random sentences drawn from two novels excluded from the systems' training data:\ Virginia Woolf's \textit{Orlando} and Radclyffe Hall's \textit{The Well of Loneliness}. The models produce the same or similar outputs for a large number of sentences and generally provide very plausible annotations. Of the 100 sentences, \texttt{t5-3b} only failed to replicate two inputs, both of which were quite long, and one of the replication errors only consisted of a dropped word. 

There were some cases in which  \texttt{t5-3b} appeared to perform better than BookNLP:\ it was better at identifying when pronouns referred to objects and not people (Row 1, Table \ref{tab:bookNLPComp}), it sometimes identified the correct coreference cluster when BookNLP failed (Row 2, Table \ref{tab:bookNLPComp}), and in one interesting case it was able to correctly cluster a name and pronouns despite a disconnect between the expected gender of the name and the gender of the pronouns (Row 3, Table \ref{tab:bookNLPComp}). However, in some cases BookNLP caught edge cases that \texttt{t5-3b} did not:\ it more accurately identified entities written in dialect (Row 4, Table \ref{tab:bookNLPComp}) and it occasionally caught less explicit entities (such as `one' or `others') that the model did not (Row 5, Table \ref{tab:bookNLPComp}). 

Overall, however, the performance of the two models appeared to be largely comparable. Despite this, we still consider the \texttt{t5-3b} model's performance to be significant for two reasons. Whereas the BookNLP pipeline required extensive development and would be very labor intensive to replicate for other data genres, fine-tuning the T5 models is simple and adaptable. In addition, the BookNLP pipeline is restricted to performing the tasks for which it has currently been trained; we view this as a promising calibration for the seq2seq models' ability to perform tasks that the LSTM cannot, such as relationship identification and characterization.

\paragraph{mT5} We also tested variations of T5 trained on larger multilingual collections. 
Although it performs worse than \texttt{t5-base} and larger, \texttt{mt5-large} reaches relatively high-performance. This performance may be boosted using additional training data, thus making it a viable option for further exploration into multi-lingual coreference annotation. Currently, when fed prompting sentences in German, Spanish, and French the model is able to reproduce sentences and identify some basic entities. However, it struggles with longer sentences and more complicated or opaque entities (Table \ref{tab:multiInputOutput}).


\begin{table}[]
\small
\caption{A \texttt{mt5-large} model fine-tuned only on English has some ability to identify entities in non-English text.}
\centering
\begin{tabular}{p{0.45\linewidth} | p{0.45\linewidth}}
\hline
\textbf{Input} &  \textbf{Output} \\
\hline
-La condesa de Albornoz -respondió el niño. & [La condesa de Albornoz: 1] -respondió [el niño: 2].\\
\hline
Mes parents ne peuvent plus faire autrement. & [Mes: 1] parents ne peuvent plus faire autrement. \\
\hline 
Und vor ihm, in der Ferne da drüben, stiegen die blauen Bergriesen auf. & And vor [ihn: 1], in der Ferne da drüben, stiegen die blauen Bergriesen auf. \\
\hline
\end{tabular}
\label{tab:multiInputOutput}
\end{table}

\paragraph{Pythia}
Previous work has only considered encoder-decoder architectures. We evaluate the open-source decoder-only Pythia model family \cite{biderman2023pythia}. 
Pythia-based models are frequently able to replicate inputs. However, they usually append extensive hallucination to the replicated input, often adding repeating substrings, brackets, or integers (Row 2, Table \ref{tab:inputOutputResults}). They very rarely include any formatting in the replicated text resembling that used for the coreference annotations. Thus, these models are currently unsuable for this task.

\section{Conclusion}

Fine-tuned generative LLMs show great promise for coreference annotation. They are simple to apply and can be efficiently trained for specific corpora from open-source base models. The errors made by large models in replicating inputs are minor and they are able to capture great complexity in the entities they annotate. In the future, we hope to extend this method to operate on longer contexts. Specifically, we propose to pre-pend all previously identified entities to each successive input. In addition, we believe that the high performance of the large, encoder-decoder models like \texttt{t5-3b} suggests that these models may be capable of performing more complex annotations, such as identifying emotional states or power dynamics between characters.

\section*{Acknowledgements}
We would like to thank Federica Bologna, Katherine Lee, Kiara Liu, Rosamond Thalken, Andrea Wang, Matthew Wilkens, and Gregory Yauney for their thoughtful feedback. This work was supported by the NEH project AI for Humanists and grew out of Mimno's experience as a Visiting Researcher with Michael Collins' group at Google Research.

\bibliography{anthology,eacl2023}
\bibliographystyle{acl_natbib}
\bigskip
\bigskip

\newpage

\appendix

\section{Fine-tuning Parameters}
\label{sec:appendixFine}

The fine-tuning parameters for each model can be found below. The batch size varied based on model. \\

\centering
\begin{tabular}{lc}
\hline
\textbf{Parameter} &  \textbf{Value} \\
\hline
evaluation\_strategy & epoch\\
learning\_rate & 2e-5 \\
weight\_decay & 0.01 \\
save\_total\_limit & 3 \\
num\_train\_epochs & 10 \\
gradient\_checkpointing & True \\
\hline
\end{tabular}
\label{tab:fineParam}

\end{document}